\pdfoutput=1

\documentclass[11pt]{article}

\usepackage{acl}

\usepackage{times}
\usepackage{latexsym}
\usepackage{amsmath}

\usepackage{graphicx}
\graphicspath{ {./} }

\usepackage[T1]{fontenc}

\usepackage[utf8]{inputenc}

\usepackage{microtype}

\title{Improve Discourse Dependency Parsing with Contextualized Representations}
\author{Yifei Zhou \\
  Cornell University \\
  Department of Computer Science\\
  \texttt{yz639@cornell.edu} \\\And
  Yansong Feng* \\
  WICT, Peking University \\
  MOE Key Lab. of Computational Linguistics\\
  \texttt{fengyansong@pku.edu.cn} \\}

\begin{document}
\maketitle
\begin{abstract}
Recent works show that discourse analysis benefits from modeling  intra- and inter-sentential levels separately, where proper representations for text units of different granularities are desired to capture both the meaning of text units and their relations to the context. In this paper, we propose to take advantage of transformers to encode  contextualized representations 
of units of different levels 
to dynamically capture the information required for discourse dependency analysis on intra- and inter-sentential levels. 
Motivated by the observation of writing patterns commonly shared across articles,
we propose a novel method that treats discourse relation identification as a sequence labelling task, which takes advantage of structural information from the context of extracted discourse trees, and  substantially outperforms traditional direct-classification methods. Experiments show that our model achieves state-of-the-art results on both English and Chinese datasets. Our code is publicly available\footnote{ https://github.com/YifeiZhou02/Improve-Discourse-Dependency-Parsing-with-Contextualized-Representations}.
\end{abstract}

\section{Introduction}
Discourse dependency parsing (DDP)
is the task of identifying the structure and relationship between Elementary Discourse Units (EDUs) in a document. It is a fundamental task of natural language understanding and can benefit many downstream applications, such as dialogue understanding \citep{perret-etal-2016-integer, ijcai2018-612} and question answering \citep{Ferrucci_Brown_Chu-Carroll_Fan_Gondek_Kalyanpur_Lally_Murdock_Nyberg_Prager_Schlaefer_Welty_2010, 10.1145/1277741.1277883}.


Although existing works have achieved much progress using transition-based systems~\citep{jia-etal-2018-modeling,10.1145/3152537, hung-etal-2020-complete} or graph-based models \citep{li-etal-2014-text, DBLP:journals/corr/abs-1812-00176, afantenos-etal-2015-discourse}, 
this task still remains a challenge. Different from syntactic parsing, the basic components in a discourse are EDUs, sequences of words, which are not trivial to represent in a straightforward way like word embeddings. Predicting the dependency and relationship between EDUs sometimes necessitates the help of a global understanding of the context so that contextualized EDU representations in the discourse are needed. Furthermore, previous studies have shown the benefit of breaking discourse analysis into intra- and inter-sentential levels \citep{wang-etal-2017-two}, building sub-trees for each sentence first and then assembling sub-trees to form a complete discourse tree. In this Sentence-First (Sent-First) framework, it is even more crucial to produce appropriate contextualized representations for text units when analyzing in intra- or inter-sentential levels. 

\begin{figure}[!ht]
\centering
\includegraphics[width = \linewidth]{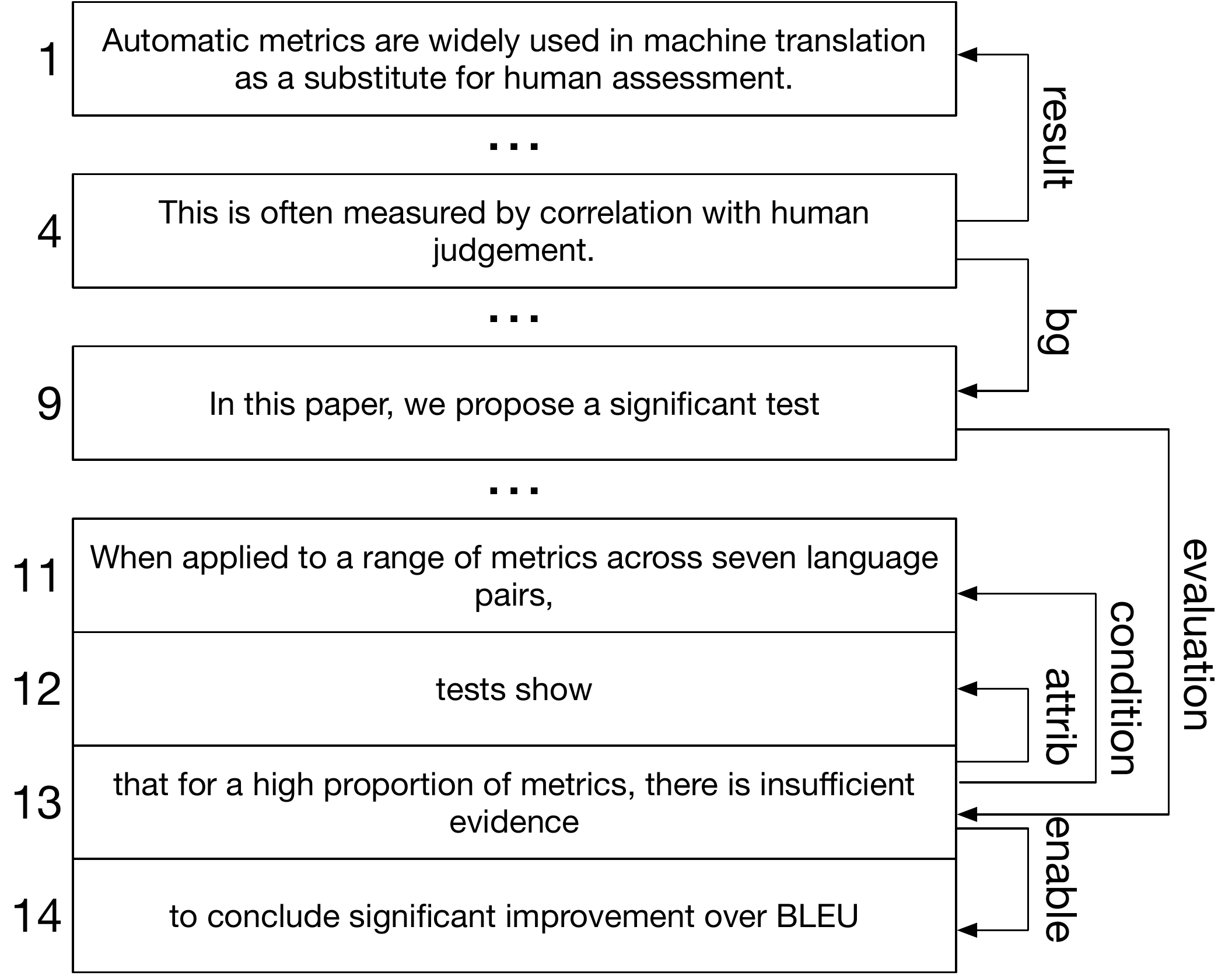}
\caption{An excerpt discourse dependency tree in SciDTB. Each indexed block is an EDU, and the origin of the arrow pointing to a particular EDU is its head. 
}
\label{fig: Introduction example}
\end{figure}

Figure 1 shows an excerpt discourse dependency structure for a scientific abstract from SciDTB \citep{yang-li-2018-scidtb}. 
The lengths of EDUs vary a lot, from more than 10 words to 2 words only (EDU 12: \textit{tests show}), making it especially hard to encode by themselves alone. Sometimes it is sufficient to consider the contextual information in a small range as in the case of EDU 13 and 14, other times we need to see a larger context as in the case of EDU 1 and 4, crossing several sentences. This again motivates us to consider encoding contextual representations of EDUs separately on intra- and inter-sentential levels to dynamically capture specific features needed for discourse analysis on different levels.

Another motivation from this example is the discovery that 
the distribution of discourse relations between EDUs seems to follow certain patterns shared across different articles. Writing patterns are document structures people commonly use to organize their arguments. For example, in scientific abstracts like the instance in Figure~1, people usually first talk about background information, then introduce the topic sentence, and conclude with elaborations or evaluations. Here, the example first states the background of widely used automatic metrics, introduces the topic sentence about their contribution of a significance test followed by evaluation and conclusion. Taking advantage of those writing patterns should enable us to better capture the interplay between individual EDUs with the context.


In this paper, 
we explore different contextualized representations for DDP in a Sent-First parsing framework, where a complete discourse tree is built up sentence by sentence. We seek to dynamically capture what is crucial 
for DDP at different text granularity levels. 
We further propose a novel discourse relation identification method that addresses the task in  a sequence labeling paradigm to exploit common conventions people usually adopt to develop their arguments. We evaluate our models on both English and Chinese  datasets, and experiments show our models achieve the state-of-the-art results by explicitly exploiting structural information in the context and capturing writing patterns that people use to organize discourses.

In summary, our contributions are mainly twofold:
(1) We incorporate the Pre-training and Fine-tuning framework into  our design of a Sent-First model and develop better contextualized EDU representations to dynamically capture different information needed for DDP at different text granularity levels. Experiments show that our model outperforms all existing models by a large margin.
(2) We formulate discourse relation identification in a novel sequence labeling paradigm to take advantage of the inherent structural information in the discourse. Building upon a stacked BiLSTM architecture, our model brings a new state-of-the-art performance on two benchmarks, showing the advantage of sequence labeling over the common practice of direct classification for discourse relation identification.

\section{Related Works}
A key finding in previous studies in discourse analysis is that most sentences have an independent well-formed sub-tree in the full document-level discourse tree \citep{joty-etal-2012-novel}. Researchers have taken advantage of this finding to build parsers that utilize different granularity levels of the document to achieve the state-of-the-art results \citep{Kobayashi_Hirao_Kamigaito_Okumura_Nagata_2020}. This design has been empirically verified to be a generally advantageous framework, improving not only works using traditional feature engineering \citep{joty-etal-2013-combining, wang-etal-2017-two}, but also  deep learning models \citep{jia-etal-2018-modeling,Kobayashi_Hirao_Kamigaito_Okumura_Nagata_2020}. We, therefore, introduce this design to our dependency parsing framework. Specifically, sub-trees for each sentence in a discourse are first built separately, then assembled to form a complete discourse tree.

However, our model differs from prior works in that we make a clear distinction to derive better contextualized representations of EDUs from fine-tuning BERT separately for intra- and inter-sentential levels to dynamically capture different information needed for discourse analysis at different levels. We are also the first to design stacked sequence labeling models for discourse relation identification so that its hierarchical structure can explicitly capture both intra-sentential and inter-sentential writing patterns. 

In the case of implicit relations between EDUs without clear connectives, it is crucial to introduce sequential information from the context to resolve ambiguity. \citet{feng-hirst-2014-linear} rely on linear-chain CRF with traditional feature engineering to make use of the sequential characteristics of the context for discourse constituent parsing. However, they greedily build up the discourse structure and relations from bottom up. At each timestep, they apply the CRF to obtain the locally optimized structure and relation. In this way, the model assigns relation gradually along with the construction of the parsing tree from bottom up, but only limited contextual information from the top level of the partially constructed tree can be used to predict relations. Besides, at each timestep, they sequentially assign relations to top nodes of the partial tree, without being aware that those nodes might represent different levels of discourse units (e.g. EDUs, sentences, or even paragraphs). In contrast, we explicitly train our sequence labeling models on both intra- and inter-sentential levels after a complete discourse tree is constructed so that we can infer from the whole context with a clear intention of capturing different writing patterns occurring at intra- and inter-sentential levels. 

\section{Task Definition}
We define the task of discourse dependency parsing as following: given a sequence of EDUs of length $l$, $(e_1, e_2,...,e_l)$ and a set of possible relations between EDUs $Re$, the goal is to predict another sequence of EDUs $(h_1,h_2,...,h_l)$ such that $\forall h_i, h_i \in (e_1, e_2,...,e_l)$ is the head of $e_i$ and a sequence of   relations $(r_1,r_2,...,r_l)$ such that $\forall r_i, r_i$ is the relation between tuple $(e_i, h_i)$. 

\section{Our Model}
We follow previous works \citep{wang-etal-2017-two} to cast the task of discourse dependency parsing as a composition of two separate yet related subtasks: dependency tree construction and relation identification. We design our model primarily in a two-step pipeline. We incorporate Sent-First design as our backbone (i.e. building sub-trees for each sentence and then assembling them into a complete discourse tree), and formulate discourse relation identification as a sequence labeling task on both intra- and inter-sentential levels to take advantage of the structure information in the discourse. Figure~1 shows the overview of our model.
\begin{figure*}[!ht]
\centering
\includegraphics[width = \linewidth]{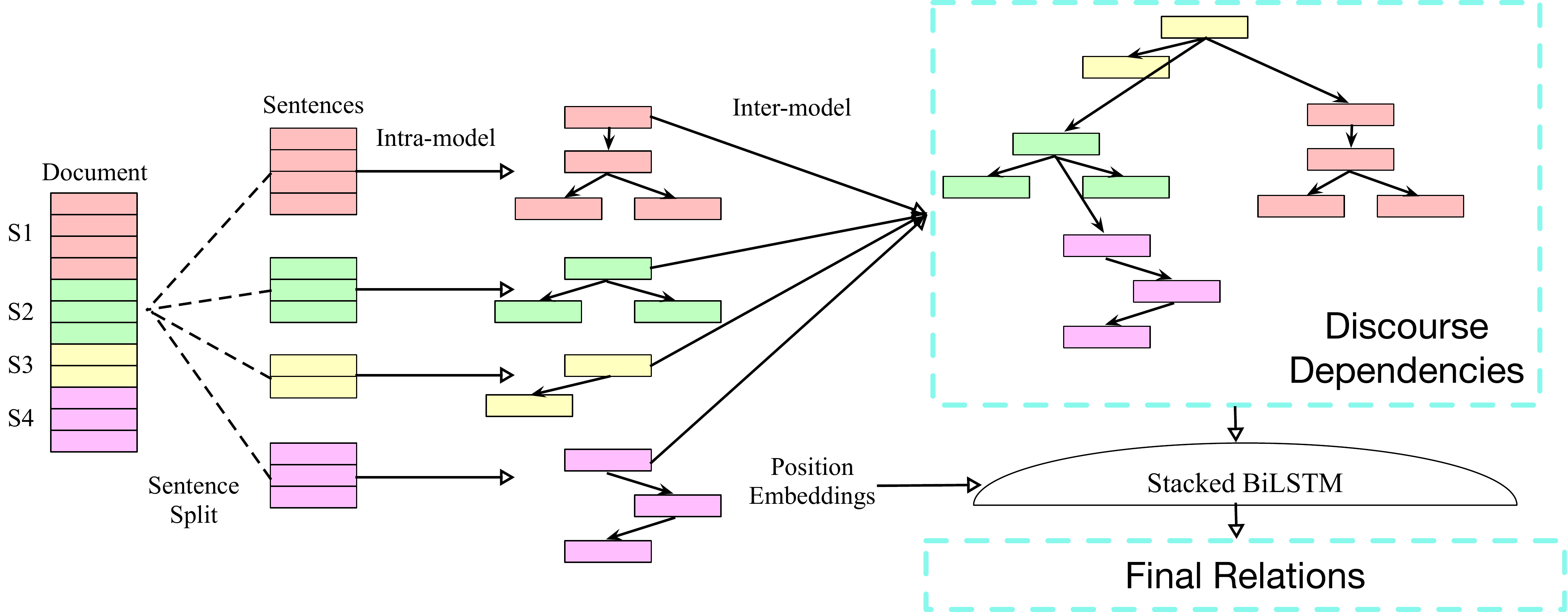}
\caption{An overview of our model. 
Intra-sentential dependencies are discovered first and inter-sentential dependencies are constructed after that to form a complete dependency tree. 
}
\label{fig:Model-Overview}
\end{figure*}

\subsection{Discourse Dependency Tree Constructor}

To take advantage of the property of well-formed sentence sub-trees inside a full discourse tree, we break the task of dependency parsing into two different levels, discovering intra-sentential sub-tree structures first and then aseembling them into a full discourse tree by identifying the inter-sentential structure of the discourse.

\paragraph{Arc-Eager Transition System} Since discourse dependency trees are primarily annotated as projective trees \citep{yang-li-2018-scidtb}, we design our tree constructor as a transition system, which converts the structure prediction process into a sequence of predicted actions. At each timestep, we derive a state feature to represent the  state, which is fed into an output layer to get the predicted action.
Our model follows the standard Arc-Eager system, with the action set: \emph{O}$ = \{Shift, Left-Arc, Right-Arc, Reduce\} $. 

Specifically, our discourse tree constructor maintains a stack \emph{S}, a queue \emph{I}, and a set of assigned arcs \emph{A} during parsing. The stack \emph{S} and the set of assigned arcs \emph{A} are initialized to be empty, while the queue \emph{I} contains all the EDUs in the input sequence. At each timestep, an action in the action set \emph{O} is performed with the following definition: \textit{Shift} pushes the first EDU in queue \emph{I} to the top of stack \emph{S}; \textit{Left-Arc} adds an arc from the first EDU in queue \emph{I} to the top EDU in stack \emph{S} (i.e. assigns the first EDU in \emph{I} to be the head of the top EDU in \emph{S}) and removes the top EDU in \emph{S}; \textit{Right-Arc} adds an arc from the top EDU in stack \emph{S} to the first EDU in queue \emph{I} (i.e. assigns the top EDU in \emph{S} to be the head) and pushes the first EDU in \emph{I} to stack \emph{S}; \textit{Reduce} removes the top EDU in \emph{S}. Parsing terminates when \emph{I} becomes empty and the only EDU left in \emph{S} is selected to be the head of the input sequence. More details of Arc-Eager transition system can be referred from \citet{nivre-2003-efficient}.

We first construct a dependency sub-tree for each sentence, and then treat each sub-tree as a leaf node to form a complete discourse tree across sentences. In this way, we can break a long discourse into smaller sub-structures to reduce the search space. A mathematical bound for the reduction of search space of our Sent-First framework for DDP and discourse constituent parsing is also provided in Appendix.


 \paragraph{Contextualized State Representation}
Ideally, we would like the feature representation to contain both the information of the EDUs directly involved in the action to be executed and rich clues from the context from both the tree-structure and the text, e.g. the parsing history and the interactions between individual EDUs in the context with an appropriate scope of text. In order to capture the structural clues from the context, we incorporate the parsing history in the form of identified dependencies in addition to traditional state representations to represent the current state. At each timestep, we select 6 EDUs from the current state as our feature template, including the first and the second EDU at the top of stack \emph{S}, the first and the second EDU in queue \emph{I}, and the head EDUs for the first and the second EDU at the top of stack \emph{S}, respectively. A feature vector of all zeros is used if there is no EDU at a certain position.

\paragraph{EDU Representations}
To better capture an EDU in our Sent-First framework, we use pre-trained BERT \citep{DBLP:journals/corr/abs-1810-04805} to obtain representations for each EDU according to different context. We argue that an EDU should have different representations when it is considered in different parsing levels, and thus requires level-specific contextual representations.
For intra-sentential tree constructor, we feed the entire sentence to BERT and represent each EDU by averaging the last hidden states of all tokens in that EDU. The reason behind is that sentences are often self-contained sub-units of the discourse, and it is sufficient to consider interactions among EDUs within a sentence for intra-sentential analysis.
On the other hand, 
for inter-sentential tree constructor, we concatenate all the root EDUs of different sentences in the discourse to form a pseudo sentence, feed it to BERT, and similarly, represent each root EDU by averaging the last hidden states of all tokens in each root EDU. In this way, we aim to encourage EDUs across different sentences to directly interact with each other, in order to reflect the global properties of a discourse.
Figure 2 shows the architecture for our two-stage discourse dependency tree constructor. 


\subsection{Discourse Relation Identification}

After the tree constructor is trained, we train separate sequence labeling models for relation identification. Although discourse relation identification in discourse dependency parsing is traditionally treated as a classification task, where the common practice is to use feature engineering or neural language models to directly compare two EDUs involved isolated from the rest of the context \citep{li-etal-2014-text, DBLP:journals/corr/abs-1812-00176, yi-etal-2021-unifying}, sometimes relations between EDU pairs can be hard to be classified in isolation, as global information from the context like how EDUs are organized to support the claim in the discourse is sometimes required to infer the implicit discourse relations without explicit connectives. Therefore, we propose to identify discourse relation identification as a sequence labeling task.

\paragraph{Structure-aware Representations}
For sequence labeling, we need proper representations for EDU pairs to reflect the structure of the dependency tree. Therefore, we first tile each EDU in the input sequence $(e_1, e_2,...,e_l)$ with their predicted heads to form a sequence of EDU pairs $((e_1, h_1), (e_2, h_2),...,(e_l, h_l))$. Each EDU pair is reordered so that two arguments appear in the same order as they appear in the discourse. We derive a relation representation for each EDU pair with a BERT fine-tuned on the task of direct relation classification of EDU pairs with the [CLS] representation of the concatenation of two sentences. 

\paragraph{Position Embeddings}
We further introduce position embeddings for each EDU pair $(e_i, h_i)$, where we consider the position of $e_i$ in its corresponding sentence, and the position of its sentence in the discourse.  Specifically, we use cosine and sine functions of different frequencies~\citep{DBLP:journals/corr/VaswaniSPUJGKP17} to include position information as:
\begin{align*}
    PE_j = sin(No/10000^{j/d})+ cos(ID/10000^{j/d})
\end{align*}
where $PE$ is the position embeddings, $No$ is the position of the sentence containing $e_i$ in the discourse, $ID$ is the position of $e_i$ in the sentence, \emph{j} is the dimension of the position embeddings, $d$ is the dimension of the relation representation. The position embeddings have the same dimension as relation representations, so that they can be added directly to get the integrated representation for each EDU pair.

\paragraph{Stacked BiLSTM} 
We propose a stacked BiLSTM neural network architecture to capture both intra-sentential and inter-sentential interplay of EDUs. After labeling the entire sequence of EDU pairs $((e_1, h_1), (e_2, h_2),...,(e_l, h_l))$ with the first layer of BiLSTM, we select the root EDU for each sentence (namely the root EDU selected from our intra-sentential tree constructor for each setence) to form another inter-sentential sequence. Another separately trained BiLSTM is then applied to label those relations that span across sentences. Note that we will overwrite predictions of inter-sentential relations of the previous layer if there is a conflict of predictions.

\begin{figure*}[!ht]
\centering
\includegraphics[width = \linewidth]{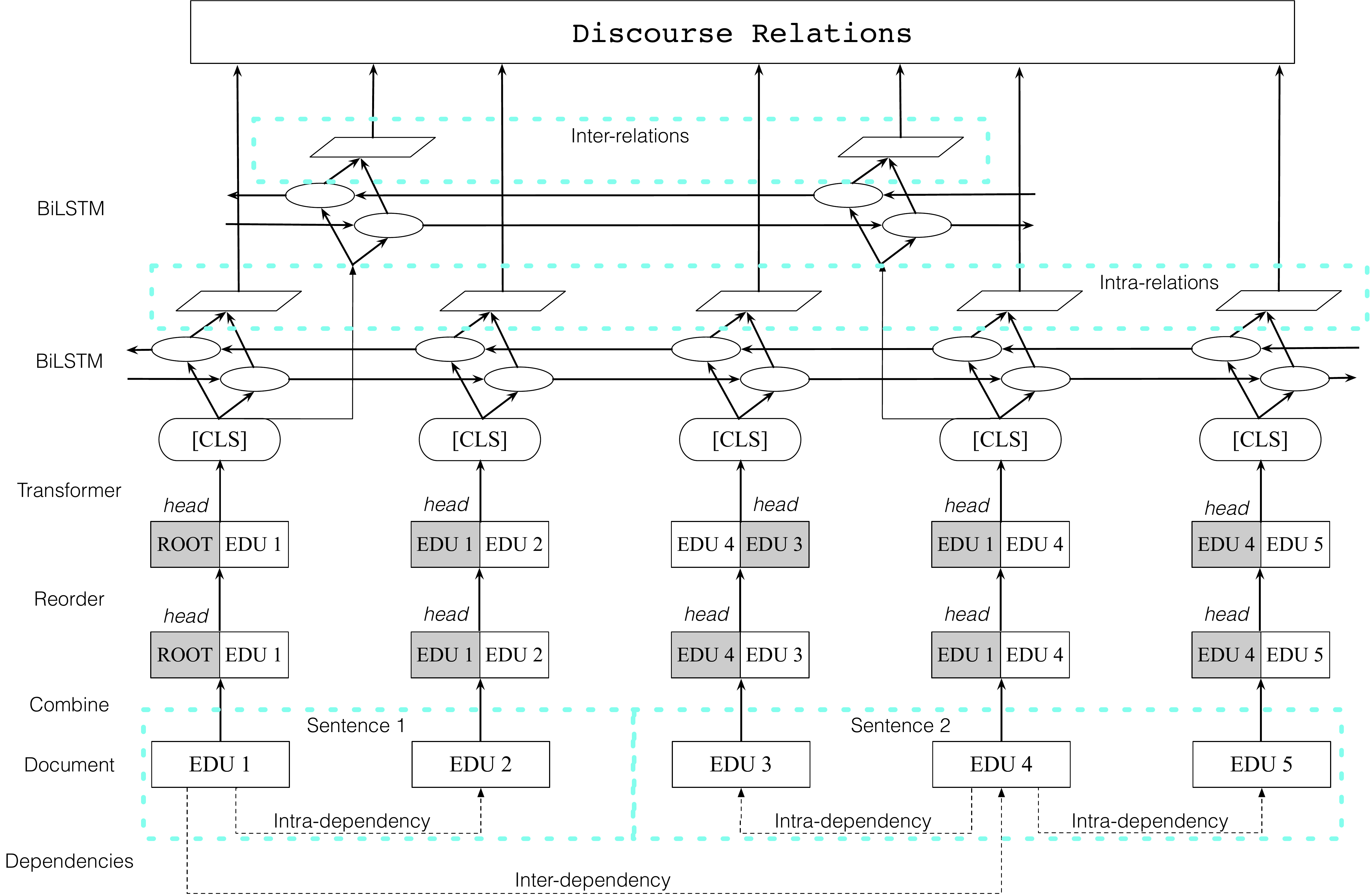}
\caption{The architecture of our relation labeling stacked BiLSTM model. Hierarchical sequence labeling is used for labeling relations on intra-sentential and inter-sentential levels.}
\label{fig: Relation Sequence Tagging}
\end{figure*}

\subsection{Training}
Our models are trained with offline learning. We train the tree constructor and the relation labeling models separately. We attain the static oracle to train tree constructors and use the gold dependency structure to train our discourse relation labelling models. Intra- and inter-sentential tree constructors are trained separately. To label discourse relations, we fine-tune the BERT used to encode the EDU pair with an additional output layer for direct relation classification. Sequence labeling models for relation identification are trained on top of the fine-tuned BERT. We use cross entropy loss for training.

\section{Experiments}
Our experiments are designed to investigate how we can better explore contextual representations to improve discourse dependency parsing. 

We evaluate our models on two manually labeled discourse treebanks of different language, 
i.e., Discourse Dependency Treebank for Scientific Abstracts (SciDTB) \citep{yang-li-2018-scidtb} in English and Chinese Discourse Treebank (CDTB) \citep{li-etal-2014-building}. 
SciDTB contains 1,355 English scientific abstracts collected from ACL Anthology.
Averagely, an abstract includes 5.3 sentences, 14.1 EDUs, where an EDU has 10.3 tokens in average.
On the other hand, CDTB was originally annotated as connective-driven constituent trees, and manually converted into a dependency style by \citet{yi-etal-2021-unifying}. CDTB contains 2,332 news documents. 
The average length of a paragraph is 2.1 sentences, 4.5 EDUs. And an EDU contains 23.3 tokens in average. 

We evaluate model performance 
using Unlabeled Attachment Score (UAS) and Labeled Attachment Score (LAS) for dependency prediction and discourse relation identification. UAS is defined as the percentage of nodes with correctly predicted heads, while LAS is defined as the percentage of nodes with both correctly predicted heads and correctly predicted relations to their heads. We report LAS against both gold dependencies and model predicted dependencies. We adopt the fine-granularity discourse relation annotations in the original datasets, 26 relations for SciDTB and 17 relations for CDTB. 

For both datasets, we trained our dependency tree constructors with an Adam optimizer with learning rate 2e-5 for 3 epochs. Our relation labeling models are all trained with an Adam optimizer until convergence. Learning rate is set to one of \{1e-5, 2e-5, 4e-5\}. 

\subsection{Baselines}
 \paragraph{Structure Prediction}
We compare with the following competitive methods for structure prediction.
(1) \textbf{Graph} adopts the Eisner's algorithm to predict the most probable dependency tree structure \citep{li-etal-2014-text, yang-li-2018-scidtb, yi-etal-2021-unifying}. (2) \textbf{Two-stage}, which is the state-of-the-art model on CDTB and SciDTB, uses an SVM to construct a dependency tree~
\citep{yang-li-2018-scidtb, yi-etal-2021-unifying}. (3) \textbf{Sent-First LSTM} is our implmentation of the state-of-the-art transition-based discourse constituent parser on RST \citep{Kobayashi_Hirao_Kamigaito_Okumura_Nagata_2020}, where we use a vanilla transition system with pretrained BiLSTM as the EDU encoder within the Sent-First framework to construct dependency trees. (4) \textbf{Complete Parser} is modified from a state-of-the-art constituent discourse parser on CDTB \citep{hung-etal-2020-complete}, using a transition system with BERT as the EDU encoder to construct a dependency tree. Because of the inherent difference between constituency parsing and dependency parsing, we only adopt the encoding strategy of (4) and (5) into our arc-eager transition system.

We also implement several model variants for comparison and ablation study. (5) \textbf{Complete Parser (contextualized)} is our modified version of Complete Parser where, instead of encoding each EDU separately, we obtain the EDU representations by encoding the whole sentence with BERT and average the corresponding token representations for the EDU. (6) \textbf{BERT + Sent-First (shared)} incorporate different contextualized embeddings from BERT into the Sent-First framework for parsing at intra- and inter-sentential levels, with the same BERT layer shared across intra-sentential and inter-sentential parsing.  (7) \textbf{BERT + Sent-First} fine-tunes separate BERT layers for intra-sentential and inter-sentential parsing independently. 


\begin{table}[!ht]
\centering
\begin{tabular}{lccr}
 & \small SciDTB &\small CDTB \\
\textbf{\small Model} & \multicolumn{2}{c}{\small UAS} \\
\hline
\small Graph (Cheng21)    & 57.6 & 58.5  \\
\small Two-stage (Cheng21)   & 70.2 & 80.3\\
\small Sent-First LSTM (Kobayashi20) & 63.9 & / \\
\small Complete Parser (Hung20)   & 75.4 & 77.7\\
\small Complete Parser (contextualized)   & 76.1 & 79.1\\
\small \textbf{BERT + Sent-First (shared)} & 77.3   & 81.5\\
\small \textbf{BERT + Sent-First}& \textbf{79.3}  & \textbf{82.2}\\
\hline
\small \textbf{Human}  & 80.2 & 89.7
\end{tabular}
\caption{Model performance of structure prediction on  SciDTB and CDTB.}
\end{table}

\paragraph{Relation Identification}
(1) \textbf{Graph} uses an averaged perceptron to classify relations by direct classification \citep{yi-etal-2021-unifying, yang-li-2018-scidtb}. (2) \textbf{Two-stage} exploits careful feature engineering and trains an SVM to classify the relations for pairs of EDUs \citep{yi-etal-2021-unifying, yang-li-2018-scidtb}. (3) \textbf{Sent-First LSTM} uses biLSTM to encode each EDU separately and a feed forward neural network for direct relation classification. (4) \textbf{BERT} is our implementation of the state-of-the-art model from \citet{yi-etal-2021-unifying} and \citet{hung-etal-2020-complete}, which fine-tunes a BERT model with an additional output layer to directly classify both intra-sentential and inter-sentential relations. (5) \textbf{BERT + BiL} formulates dependency discourse relation identification as a sequence labeling task, training an additional layer of BiLSTM on top of the 
BERT layer finetuned on direct classification. (6) \textbf{BERT SBiL} trains another BiLSTM to label inter-sentential relations on top of the original model BERT + BiL.


\begin{table}
\centering
\begin{tabular}{lccccr}
& \multicolumn{2}{c}{\small SciDTB} & \multicolumn{2}{c}{\small CDTB} \\
\small Model & \small Gold & \small Pred. & \small Gold & \small Pred.\\
\hline
\small Graph (Cheng21)  & \small / & \small 42.5 & \small/ & \small41.5 \\
\small Two-stage (Cheng21)  & \small/ & \small54.5 & \small/ & \small58.7\\
\scriptsize Sent-First LSTM {\tiny(Kobayashi20)} & \small52.5 & \small44.6 & \small/ & \small/ \\
\small BERT (Cheng21)   & \small75.5 & \small63.6 & \small74.9 & \small64.1\\
\small \textbf{BERT + BiL}  & \small76.6 & \small64.8 & \small\textbf{76.5} & \small\textbf{64.8}\\
\small \textbf{BERT + SBiL}  &  \small\textbf{77.4}& \small\textbf{65.0} & \small\textbf{76.5} & \small64.4\\
\hline
\small \textbf{Human}   & \small / & \small 62.2 & \small / & \small77.4
\end{tabular}
\caption{Model performance of relation identification on SciDTB and CDTB.}
\end{table}

\subsection{Main Results}
\paragraph{Dependency Prediction}
Table 1 summarizes the performances of different models on both datasets in terms of UAS. For traditional feature engineering models, Two-stage has already achieved satisfactory performance, even beating several neural models like Sent-First LSTM and Complete Parser. This is probably because traditional feature engineering methods design delicate structural features in addition to representations of EDUs so that they can include contextual clues to facilitate parsing.  Complete Parser leverages the benefit of better representations from pre-trained transformers to encode the information of individual EDUs, achieving a significant improvement over Sent-First LSTM model with LSTM as primary encoders. However, we show that our model BERT + Sent-First that exploits the potential of Sent-First framework with proper contextualized representations to capture the interactions between individual EDUs and the context surpasses all the existing baselines. The performance of our model can be further improved if we encode contextualized embeddings separately for intra-sentential and inter-sentential parsing to dynamically capture different information required to parsing at different text granularity levels.

\paragraph{Relation Identification}

Although previous methods like Graph, Two-stage, and Sent-First LSTM achieve decent results on both datasets, their performances are not comparable to transformer methods developed in recent years. BERT (Cheng21) is our implementation of the state-of-the-art method for relation classification in discourse dependency parsing, which improves the baseline by a large margin. Although BERT is still a very strong baseline in many NLP tasks, direct classification with BERT neglects the contextual clues in the discourse that can be exploited to aid discourse relation identification, as have been discussed in section 1. We show that the results can be further improved by making use of the sequential structure of the discourse. We design multiple novel sequence labeling models on top of the fine-tuned BERT and all of them achieve a considerable improvement (more than 1\%) over BERT in terms of accuracy both on the gold dependencies and the predicted dependencies from our Sent-First (separate), showing the benefit of enhancing the interactions between individual EDUs with the context. It yields another large gain when we introduce another layer of inter-sentential level BiLSTM, showing again that it is crucial to capture the interactions between EDUs and their context in both intra- and inter-sentential levels.

\subsection{Detailed Analysis}

 
\paragraph{Contextualized Representations for Tree Construction}

\begin{table}
\centering
\begin{tabular}{lccccr}
&\multicolumn{2}{c}{\small SciDTB} & \multicolumn{2}{c}{\small CDTB} \\
\small Model & \small intra- & \small inter- & \small intra- & \small inter- \\
\hline
\tiny Complete Parser (contextualized)& \small85.6 & \small60.7  & \small79.9  & \small78.0\\
\scriptsize BERT+Sent-First (shared)  &\small87.6 & \small 61.1 & \small81.5 & \small81.6\\
\scriptsize BERT+Sent-First  & \small \textbf{88.5} & \small\textbf{64.7} & \small\textbf{82.5} & \small\textbf{82.0}\\
\end{tabular}
\caption{Model performance (UAS) on intra- and inter-sentential dependencies. }
\end{table}
Intuitively, a model should take different views of context when analyzing intra- and inter-sentential structures. As we can see in Table~1, BERT + Sent-First (shared) improves Complete Parser (contextualized) by 1.2\% and 2.4\% on SciTDB and CDTB, respectively. The only difference is BERT + Sent-First makes explicit predictions on two different levels, while Complete Parser (contextualized) treats them equally. When we force BERT + Sent-First to use different BERTs for intra- and inter-sententential analysis, we observe further improvement, around 3\% on both datasets.

If we take a closer look at their performance in intra- and inter-sentential views in Table~3, we can see that  BERT + Sent-First (shared) performs better than single BERT model, Complete Parser (contextualized), on both intra- and inter- levels of SciDTB and CDTB, though in some cases we only observe marginal improvement like inter-sentential level of SciDTB. However, when we enhance BERT + Sent-First with different encoders for intra- and inter-sentential analysis, we can observe significant improvement in all cases. That again shows the importance of anaylzing with different but more focused contextual representations for the two parsing levels.  




\paragraph{Classification or Sequence Labeling?}

\begin{table}
\centering
\begin{tabular}{lccc}
 & \small BERT & \small BERT+BiL & \small BERT+SBiL\\
\hline
intra- & 81.8 & \textbf{82.4} & \textbf{82.4} \\
inter- & 58.1 & 60.2 & \textbf{62.6} \\
\end{tabular}
\caption{Model performance (classification accuracy) on intra- and inter-sentential relations on SciDTB with gold dependencies. 'ROOT' relation is not counted. }
\end{table}

\begin{table}
\centering
\begin{tabular}{lccc}
 & \small BERT & \small BERT+BiL & \small BERT+SBiL\\
\hline
original & 72.0 & 71.8 & \textbf{73.6} \\
modified & 50.9 & 52.3 & \textbf{53.4} \\
\end{tabular}
\caption{Model performance (classification accuracy) on automatically generated implicit relation extraction on SciDTB before and after modification.}
\end{table}

Most previous works treat discourse relation identification as a straightforward classification task, where given two EDUs, a system should identify which relationship the EDU pair hold. 
As can be seen from Table 2, all sequence labeling models (our main model as well as the variants) achieve a considerable gain over direct classification models on both datasets, especially in terms of accuracy on gold dependencies. This result verifies our hypothesis about the structural patterns of discourse relations shared across different articles. It is noticed that BERT + SBiL performs the best because its hierarchical structure can better capture different structured representations occuring at intra- and inter-sentential levels. 

In Table 4, we include the performances of different models on intra- and inter-sentential relations on SciDTB with gold dependency structure. We observe that although our BERT+BiL model improves accuracies on both levels compared to the traditional classification model, the more significant improvement is on the inter-sentential level (by 2.1\%). We show that it can even be promoted by another 2.4\% if we stack an additional BiLSTM layer on top to explicitly capture the interplay between EDUs on the inter-sentential level. That's probably because writing patterns are more likely to appear in a global view so that discourse relations on the inter-sentential level tend to be more structurally organized than that on the intra-sentential level. 

To test the effectiveness of our model for implicit discourse relation identification, We delete some freely omissible connectives identified by \citet{ma-etal-2019-implicit} to automatically generate implicit discourse relations. This results in 564 implicit instances in the test discourses. We run our model on the modified test data without retraining and compare the accuracies on those generated implicit relations. Table 5 shows the accuracies for those 564 instances before and after the modification. After the modification, although accuracies of all three models drop significantly, our sequence labeling model BERT+BiL and BERT+SBiL outperform the traditional direct classification model BERT by 1.4\% and 2.5\% respectively, showing that our sequence labeling models can make use of clues from the context to help identify relations in the case of implicit relations.

In addition, we experiment with other empirical implementations of contextualized representations instead of averaging tokens like using [CLS] for aggregate representations of sentences for inter-sentential dependency parsing, but we did not observe a significant difference. Averaging token representations turns out to have better generalizability and more straightforward for implementation.

\subsection{Case Study}
For the example shown in Figure~1, the relation between EDU 9 and EDU 13 is hard to classify using traditional direct classification because both of them contain only partial information of the sentences but their relation spans across sentences. Therefore, traditional direct classification model gets confused on this EDU pair and predicts the relation to be "elab-addition", which is plausible if we only look at those two EDUs isolated from the context. However, given the gold dependency structure, our sequence labeling model fits the EDU pair into the context and infers from common writing patterns to successfully yield the right prediction "evaluation". This shows that our model can refer to the structural information in the context to help make better predictions of relation labels.

\section{Conclusion}
In this paper, we incorporate  contextualized representations to our Sent-First general design of the model to dynamically capture different information required for discourse analysis on intra- and inter-sentential levels. We raise the awareness of taking advantage of writing patterns in discourse parsing and contrive a paradigm shift from direct classification to sequence labeling for discourse relation identification. We come up with a stacked biLSTM architecture to exploit its hierarchical design to capture structural information occurring at both intra- and inter-sentential levels. Future work will involve making better use of the structural information instead of applying simple sequence labeling.

\section*{Acknowledgements}
This work is supported in part by 
NSFC (62161160339). We would like to thank the anonymous reviewers and action editors for their helpful comments and suggestions. 

\bibliographystyle{acl_natbib}
\bibliography{anthology,custom}
\appendix
\section{Proof of Theorems}
\textbf{\textit{Theorem 1:}} For a document $D$ with $m$ sentences $(s_1, s_2,...,s_m)$ and $n$ of the sentences have length(in terms of the number of EDUs) greater or equal to 2 satisfying $|s_i| \ge 2$. Let $T$ be the set of all projective dependency trees obtainable from $D$, and let $T'$ be the set of all projective dependency trees obtainable from $D$ in a \textit{Sent-First} fashion. Then the following inequality holds: 
\begin{align*}
    |T'| &\le \frac{2}{n+1}|T|
\end{align*}

\paragraph{Proof of Theorem 1: }By the definition of our \textit{Sent-First}  method, trees in $T'$ satisfy the property that there is exactly one EDU in each sentence whose head or children lies outside the sentence. It is clear that $T' \subset T$. We consider a document $D$ with $m$ sentences $(s_1, s_2,...,s_m)$ and $n$ of the sentences have length(in terms of the number of EDUs) greater or equal to 2 satisfying $|s_i| \ge 2$. 

$\forall \sigma' \in T'$, $\sigma'$ is a valid projective dependency tree obtainable from $D$ in a \textit{Sent-First} fashion. We define a $t$-transformation to a sentence $s_i, |s_i| > 1$ with its local root of the sentence $e_{ia}$ not being the root of the document in $\sigma'$ with the following rules:
\begin{enumerate}
    \item If $e_{ia}$ has no child outside $s_i$, $e_{ib}$ is its furthest (in terms of distance to $e_{ia}$) child or one of its furthest children inside $s_i$, then delete the edge between $e_{ia}$) and $e_{ib}$ and set the head of $e_{ib}$ to be the head of $e_{ia}$.
    \item Else if $e_{ia}$ has at least one child before $e_{ia}$ inside $s_i$, and $e_{ib}$ is its furthest child before $e_{ia}$ inside $s_i$. Delete the edge between $e_{ia}$ and $e_{ib}$. If $i > 1$, set the head of $e_{ib}$ to be the local root of sentence $s_{i-1}$, else $i = 1$, set the head of $e_{ib}$ to be the local root of sentence $s_{i+1}$.
    \item Else, $e_{ia}$ has at least one child after $e_{ia}$ inside $s_i$, and $e_{ib}$ is its furthest child after $e_{ia}$ inside $s_i$. Delete the edge between $e_{ia}$) and $e_{ib}$. If $i < m$, set the head of $e_{ib}$ to be the local root of sentence $s_{i+1}$, else $i = m$, set the head of $e_{ib}$ to be the local root of sentence $s_{m - 1}$.
\end{enumerate}
Suppose $\sigma_i$ is obtained by applying $t$-transformation to the sentence $s_i$, it is obvious to show that $\sigma_i \in T/T'$. $n-1$ valid $t$-transformations can be applied to $\sigma'$. A reverse transformation $t^{-1}$ can be applied to $\sigma_i$ with the following rule: if a sentence has two local roots, change the head of one of the roots to the other root. In this way, at most two possibly valid trees $\in T'$ can be obtained because we are not sure which one is the original local root of the sentence. Therefore, at most 2 different $\sigma' \in T'$ can be found to share the same tree structure after a $t$-transformation. See Figure 5 for illustration. Therefore, 
\begin{figure}[!ht]
\includegraphics[width = 0.8\linewidth]{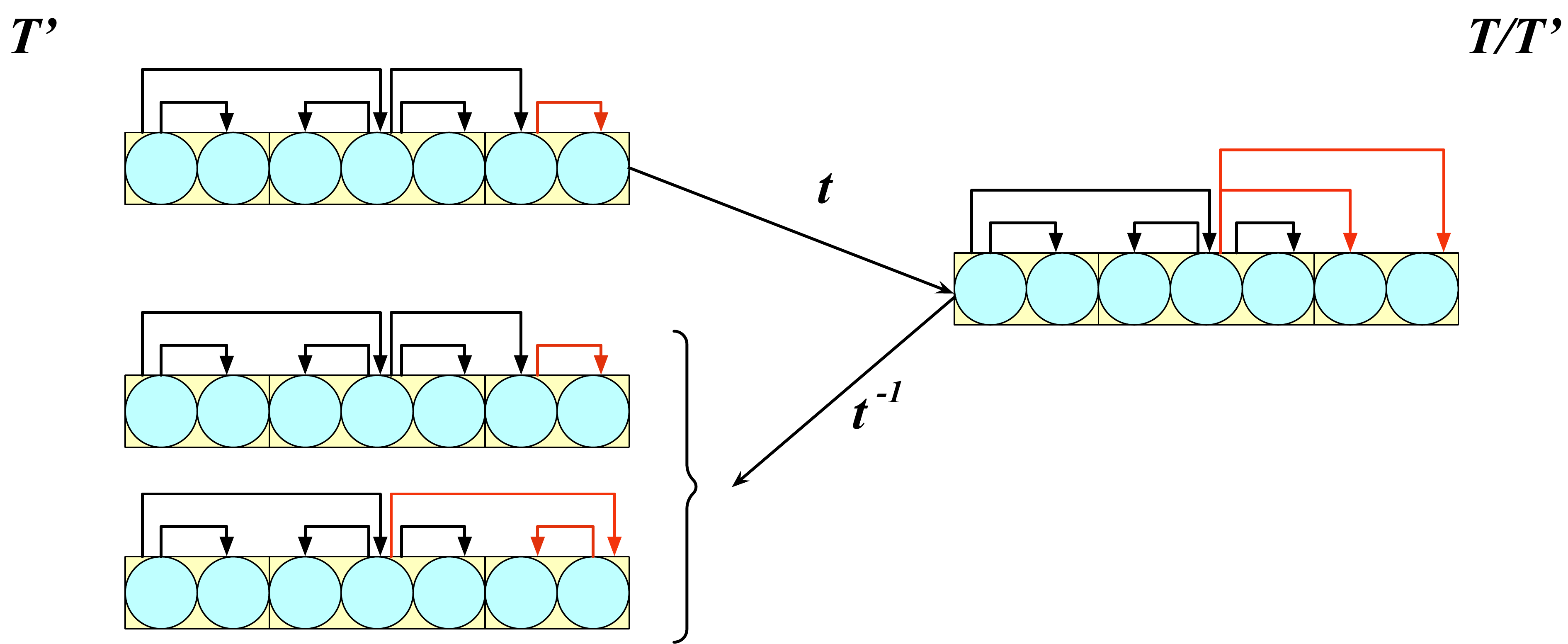}
\caption{An illustration of transformation $t$ for \textbf{\textit{Theorem 1}}.}
\end{figure}
\begin{align*}
    |T/T'| &\ge \frac{n-1}{2}|T'|\\
    |T'| &\le \frac{2}{n+1}|T|
\end{align*}

\textit{Theorem 1} shows that the search space shrinks with the number of sentences. Therefore, \textit{Sent-First} approach is especially effective at the reduction of search space so that the parser has a better chance to find the correct result, no matter what kind of parser is used specifically. Since the effectiveness has been proved, this approach can even be confidently generalized to other cases where similar sentence-like boundaries can be identified. 

Besides, an even stronger bound regarding the use of \textit{Sent-First} method can also be proved for constituent parsing.

\textbf{\textit{Theorem 2:}} For a document $D$ with $m > 1$ sentences $(s_1, s_2,...,s_m)$ and $n$ of the sentences have length(in terms of the number of EDUs) greater or equal to 2 satisfying $|s_i| \ge 2$. Let $T$ be the set of all binary constituency trees obtainable from $D$, and let $T'$ be the set of all binary constituency trees obtainable from $D$ in a \textit{Sent-First} fashion. Then the following inequality holds: 
\begin{align*}
    |T'| \le (\frac{1}{2})^{n}|T|
\end{align*}
\paragraph{Proof of Theorem 2: }By the definition of our \textit{Sent-First}  method, trees in $T'$ satisfy the property that EDUs in a sentence forms a complete sub-tree. It is clear that $T' \subset T$. We define a tree transformation $t$, for a tree $u_1$ with child $u_2$ and $u_3$, $u_3$ being a complete discourse tree of a sentence with more than 2 EDUs. $u_3$ must also have 2 children named $u_4$ and $u_5$ where $u_4$ is adjacent to $u_2$ in the sentence. After transformation $t$, a new tree $u_1'$ is derived whose children are $u_5$ and a sub-tree $u_6$ with children $u_2$ and $u_4$. $u_1 \in T'$, while $u_1' \in T/T'$. Illustration see Figure 6. Note that $t$ is one-to-one so that different $u_1$ will be transformed to different $u_1'$ after $t$-transformation and $u_1$ can be applied $t$-transformation twice if both children of $u_1$ are complete DTs for a sentence (more possible trees $u_1'$ can be transformed into if the order of transformation is also considered). Transformation $t$ is a local transformation and does not affect sub-trees $u_2$, $u_4$, and $u_5$. 

\begin{figure}[!ht]
\includegraphics[width = 0.8\linewidth]{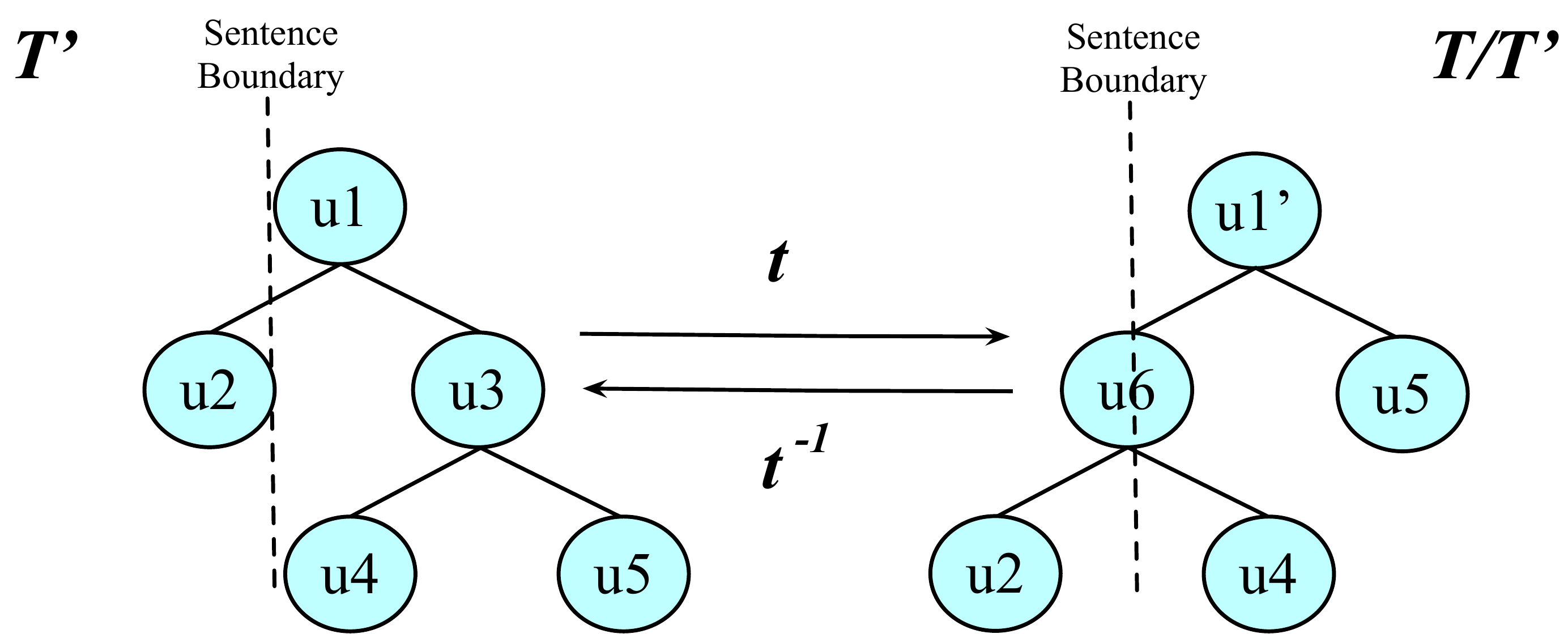}
\caption{An illustration of transformation $t$ for \textbf{\textit{Theorem 2}}.}
\end{figure}

$\forall \sigma' \in T'$, $\sigma'$ is a valid projective dependency tree obtainable from $D$ in a \textit{Sent-First} fashion. Since all sub-trees representing a sentence must merge into one complete discourse tree representing the whole document, there must be $n$ independent $t$ transformations applicable to some sub-trees in $\sigma'$, so that at least $2^n - 1$ trees can be obtained after $i \ge 1$ $t$ transformations $\in T/T'$. Since $t$-transformation is one-to-one, $\forall \sigma_1, \sigma_2 \in T', \sigma_1 \ne \sigma_2$, $ \sigma_1'$ is a tree obtained after some $t$-transformations on $ \sigma_1$, $ \sigma_2'$ is a tree obtained after some $t$-transformations on $ \sigma_2$, $\sigma_1' \ne \sigma_2'$. 

Therefore, 
\begin{align*}
    |T/T'| &\ge (2^n - 1)|T'|\\
    |T'| &\le (\frac{1}{2})^{n}|T|
\end{align*}

\section{Additional Detailed Results}
\begin{table}[!ht]
\centering
\begin{tabular}{lccc}
\tiny Relation & \tiny BERT & \tiny BERT+BiL & \tiny BERT+SBiL\\
\hline
\tiny elab-addition & 77.5& 78.9& \textbf{80.2}\\
\tiny evaluation & 76.3& 77.8& \textbf{81.6}\\
\tiny joint & 81.7& 80.4& \textbf{82.5}\\
\tiny attribution & 92.7& \textbf{95.5}& \textbf{95.5}\\
\tiny enablement & 82.1& \textbf{84.1}& 83.4\\
\tiny manner-means & \textbf{86.2} & 85.0& \textbf{86.2}\\
\tiny contrast & 73.9& 75.0& \textbf{77.1}\\
\tiny bg-goal & 59.3 & 63.5& \textbf{67.7}\\
\tiny same-unit & 89.7& \textbf{93.2}& \textbf{93.2}\\
\tiny progression & \textbf{19.0} & 6.1& 15.4\\
\tiny bg-compare & 43.8 & 44.1& \textbf{60.9} \\
\tiny elab-aspect & 29.2& 28.1& \textbf{36.2}\\
\tiny bg-general & 70.2 & \textbf{94.3}& 91.7\\
\tiny condition & \textbf{57.1} & 54.2& 52.0\\
\hline

\end{tabular}
\caption{ Model performance (F1 score) for the 14 most frequent relation types on gold dependencies of SciDTB. 
The first 14 relations are listed in descending order in terms of their frequencies in the test dataset (652, 178, 156, 131, 127, 121, 71, 56, 54, 48, 46, 45, 37, 33).
}
\end{table}

\begin{table}[!ht]
\centering
\begin{tabular}{lccc}
\tiny Span  & \tiny BERT & \tiny BERT+BiL & \tiny BERT+SBiL\\
\hline
 1 & 82.7& \textbf{83.1}& 82.9\\
 2 & 63.6& \textbf{67.5}& 67.1\\
 3 & 51.6& 55.6& \textbf{59.5}\\
 4 & \textbf{61.0}& 58.4& 59.7\\
 5 & 52.2& 53.7& \textbf{62.7}\\
 6 & \textbf{63.0} & \textbf{63.0}& 60.9\\
 7 & 70.6& \textbf{73.5}& 58.9\\
 8 & 52.9 & 50.0& \textbf{73.5}\\
 9 & 64.0& 64.0& 64.0\\
\hline

\end{tabular}
\caption{ Model performance (accuracy) of relations with gold dependencies on SciDTB against their spans. 
}
\end{table}


\end{document}